# Convolutional versus Dense Neural Networks: Comparing the Two Neural Networks' Performance in Predicting Building Operational Energy Use Based on the Building Shape

Farnaz Nazari[1], Wei Yan[1]
[1]Department of Architecture, Texas A&M University, College Station, TX, USA

## Abstract

A building's self-shading shape impacts substantially on the amount of direct sunlight received by the building and contributes significantly to building's operational energy use, in addition to other major contributing variables, such as materials and window-to-wall ratios. Deep Learning has the potential to assist designers and engineers by efficiently predicting building energy performance. This paper assesses the applicability of two different neural networks' structures, Dense Neural Network (DNN) and Convolutional Neural Network (CNN), for predicting building operational energy use with respect to building shape. The comparison between the two neural networks shows that the DNN model surpasses the CNN model in performance, simplicity, and computation time. However, image-based CNN has the benefit of utilizing architectural graphics that facilitates design communication.

## Key Innovations

- Deep Learning methods are successfully applied to predicting building energy use with synthesized training data - building shape design and simulation, which are parametrically generated.
- The comparison between the two neural networks shows that the DNN model surpasses the CNN model in performance, simplicity, and computation time. However, image-based CNN has the benefit of utilizing architectural graphics that facilitates design communication.

## Practical Implications

This study about using Deep Learning models to efficiently predict building energy use based on building shapes, with architectural design graphics as training data, can be combined with other studies in literature about other important building energy performance variables, to create a comprehensive Deep Learning model for practical design applications.

## Introduction

Previous studies have explored the relationship between the building shape and the operational energy use; however, the contribution of building shape is often in the shadow of the prominent contributors such as window-to-wall ratio (WWR) and thermal properties of the envelope. Nonetheless, a building's self-shading shape, where the amount of direct sunlight on the facade changes substantially based on the shape (Figure 1), contributes significantly to building's energy performance. The effect of different amount of direct sunlight on the envelope on the building operational energy use has been studied by many scholars through comparing the energy use associated with different building orientations (Ying and Li 2020, Hong et al. 2020, Lapisa 2019) or different shading devices (Kim et al. 2015). While direct sunlight on the envelope can be desired during cold seasons, it increases cooling loads during hot seasons. Therefore, it is essential that the building shape be optimized with respect to the total annual energy use.

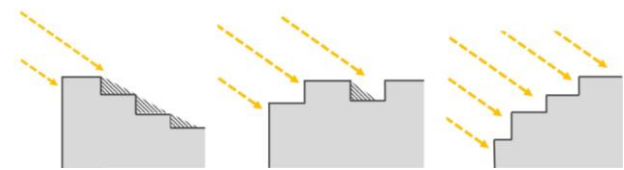

*Figure 1: The amount of direct sunlight on the building envelope varies based on the building self-shading shape*

However, building shape is not generally approached in tandem with an energy-wise design decision. One of the reasons is that making an energy-wise decision at early stages of the design is not a straightforward task since the search space is extremely large and it is not necessarily following a definable pattern. Moreover, self-shading shapes, due to having conflicting effects on building cooling, heating, and daylighting performance, cause additional complexity.

In practice, for an energy efficient building design, fundamental rules of thumb are considered, which do not suggest a particular building shape and are often limited to general recommendations on the building orientation, aspect ratio, and WWR according to the climatic zones (Olgyay 2015). In more energy-efficient oriented designs, once the documents of the schematic design phase of architecture (e.g., plans, elevations, and sections) are generated, the available tools for whole building energy simulation are employed to evaluate the energy performance of the designed model or a handful of design alternatives. However, the results do not necessarily lead to an effective conclusion as the evaluation is among a limited number of design alternatives and a benchmark model that is derived from previous experiences or codes and standards. Furthermore, given the project time and resource limitations, any changes in the building shape design, which is one of the earliest steps of the design process, would be challenging and costly. Hence, the revisions made to improve the energy efficiency of the design often ends up with modifications in façade design, construction materials, WWR, insulation specifications etc., but not the building shape.

Moreover, at the early stage of a design process there may not be enough information and details available for performing a whole building energy simulation. These complexities call for an efficient method for energy performance evaluation, which requires a minimal amount of information and details about the building design, considering time and resource limitations in the design process. Deep Learning methods that are recently employed for a broad list of applications in architectural design research have the potential to advantage designers and engineers with efficiently predicting building energy performance in early stages of the design.

This study is exploring whether the Deep Learning methods can be applicable in predicting the whole building energy use based on the building shape. To achieve this, a synthesized dataset consisting of parametrically generated building shapes and the associated building energy simulation results is used to feed the artificial neural networks as input. Rhino-Grasshopper software tools are used for creating a parametric building model to generate the building shapes. EnergyPlus is used for simulating and calculating building energy use. Two different neural networks' structures are developed and compared to assure the applicability of Deep Learning for the mentioned task. A Dense Neural Network (DNN) that employs building shape dimensions as input and a Convolutional Neural Network (CNN) that is fed with building plan images as input, are developed and compared in terms of performance, simplicity, and computation time. Noting that the possibility of an energy-wise design decision emerges at early stages of the design process, it is highly valuable that the model can employ graphical design documents as the input and be able to efficiently estimate the operational energy use based on the building shape.

## Literature Review

This section reviews the existing literature about the effect of building shape on building energy use, different building energy models, Deep Learning applications in building energy use prediction, and finally, two different neural networks' structures, DNN and CNN.

- **Building shape and building energy use**

Building shape, building geometry, and building form are often used interchangeably in the literature. Similar to the definition of shape put forward by Gatto et al. (2000) in visual arts studies as "a two-dimensional (2D) enclosed area", in this study, we refer to building footprint on the ground as building shape. Table 1 lists the studies on the relationship between building geometric aspects and building energy use, studied aspects and general terms used to describe the geometric aspects. In these studies, the relationship between building's operational energy and several building's geometric aspects such as orientation, aspect ratio, and WWR have been well documented. However, the research on the contribution of building shape, as defined in this study, on operational energy use has been strictly restricted due to certain limitations. The main limitation in studying building shape's contribution to building energy use is that the relationship between building shape and operational energy use is non-linear, and there are various variables, some of which have conflicting impacts, creating an unlimitedly broad solution pool. These limitations prevent a comprehensive analysis through the conventional methods.

*Table 1: Studies on building shape/form/geometry contribution to building operational energy use.*

| Studies | Studied aspects | General term |
|---|---|---|
| (Marks 1996, Wang et al. 2006, Adamski 2007, AlAnzi et al. 2009, Choi et al. 2012, Zhang et al. 2017, Taleb et al. 2020, Ciardiello et al. 2020, Raji et al. 2017, Gan et al. 2019) | Building shape | Building shape, Building design, Geometry parameters, Building form, Floor shape, Layout plan |
| (Ying and Li 2020) | Building shape, Perimeter-Area ratio, Orientation | Floor shape |
| (Pathirana et al. 2019) | Building shape, Orientation, WWR | Building shape, orientation, and WWR |
| (Florides et al. 2002) | Building shape, Orientation, Overhangs | Measures |
| (Ourghi et al. 2007) | Building shape, Compactness factor | Shape |
| (Premrov et al. 2018, Wei et al. 2016) | Orientation, Aspect ratio, WWR, Scale, Stacking effect | Building shape |
| (Hong et al. 2020) | Orientation, Number of floors, WWR, Plan form | Building characteristics |
| (Lapisa 2019) | Aspect ratio, Orientation | Building geometric shape and orientation |
| (Camporeale and Mercader-Moyano 2019) | Orientation, Aspect ratio, Height | Building shape |
| (Won et al. 2019) | Number of floors, Floor to area ratio, WWR, Compactness factor, Surface -Volume ratio | Morphological factors |
| (Tian et al. 2016) | Number of floors, Scale | Built form |
| (Hemsath and A. Bandhosseini 2015) | Aspect ratio with roof variability, Stacking effect | Building geometry |
| (Susorova et al. 2013) | Window orientation, WWR, Width to depth ratio (room scale) | Geometry factors |
| (Lu et al. 2017) | Aspect ratio | Building shape |
| (Page 1974, Najjar et al. 2019) | WWR | Building shape, Design factors |

Accordingly, the existing studies on the effect of building shape on energy use dealing with mentioned limitations can be classified into three categories. The first category of studies (Florides et al. 2002, Ourghi et al. 2007, AlAnzi et al. 2009, Raji et al. 2017, Zhang et al. 2017, Ying and Li 2020) performs energy simulation on hypothetical buildings with typical shapes such as rectangular, L-shape, U-shape, and H-shape. For instance, Zhang et al. (2017) have investigated energy performance for schools with typical plans (e.g. rectangular, L-shape, U-shape, courtyard), however, the contribution of building shape is integrated with contributions of other factors such as the WWR, orientation, and atrium effect. Studies in the second category compares energy performance of real buildings with different shapes. It should be noted that for studies in this category often shape is not the only variable of the study affecting the energy use of the buildings. Hence, the building shape contribution remains unclear. For instance, Choi et al. (2012) have compared energy use of 4 high-rise apartment buildings with different building shapes.

Lastly, the third category of studies (Marks 1996, Wang et al. 2006, Adamski 2007, Taleb et al. 2020) performs optimization on a building shape with respect to energy use. As an example, Taleb et al. (2020) have optimized a building with a cubic form based on the energy performance. Although the studies reviewed in these three categories address important aspects of relationship between building shape and energy use, the research that specifically focuses on the impact of building shape on energy use, as in a controlled study, is limited in the literature. Moreover, due to the computational requirements for the energy simulation and optimization processes, available studies often focus on a limited range of building shapes for the analysis, which does not allow a comprehensive analysis over a substantial number of alternative solutions.

- **Building energy models**

Common methods for building energy performance evaluation include physics-based models, data-driven models, and hybrid models (Sun et al. 2020). The physics-based models are typically simulation-based models, with which the energy-related objectives are calculated by a building performance simulation tool such as EnergyPlus, TRNSYS or ESP-r. Although simulation-based models of buildings can produce relatively accurate prediction of building energy use, the computation time of the model, and the requirement of detailed information about the design, along with the need for skilled experts to avoid garbage-in, garbage-out often become concerns. Moreover, these models are typically building-specific, which increase the project costs. Therefore, when the search space is large, the calculation and optimization process may become costly and cumbersome. Unlike physics-based models, data-driven methods require less development time and a minimal amount of information about the building characteristics (Delcroix et al. 2020). Data-driven methods range from linear regression models to Machine Learning and Deep Learning methods, and can discover statistical patterns without explicit knowledge. Due to the increasing growth in the amount of building operational datasets and big data from smart buildings, and their ability to deal with non-linearity, data-driven methods, specifically Deep Learning methods have become more attractive in recent years for tasks that exhibit significant nonlinearity, such as building energy prediction (Zhou and Zheng 2020).

- **Deep Learning application in building energy prediction**

Deep Learning methods are being increasingly employed in predicting building energy use for building energy management, HVAC optimized control, and fault diagnosis purposes. Different Deep Learning networks' structures ranging from DNN (Liu et al. 2019, Wei et al. 2019), to CNN (Amarasinghe et al. 2017, Kuo and Huang 2018), Recurrent Neural Networks (Bianchi et al. 2017), and Long-Short Term Memory (Kong et al. 2019) along with their many variants have been employed to predict electricity demand and energy consumption of buildings. In these studies, the prediction inputs typically are classified in four categories: (1) meteorological data (e.g., outdoor air dry-bulb and wet bulb temperatures, relative humidity, solar radiation), (2) time variable (hour of the day, day of the week), (3) physical properties of buildings (e.g., relative compactness, surface area, glazing area), and (4) historical data of energy consumption records of buildings. Unlike previous studies, in the present study the Deep Learning input data are building shapes and associated energy consumption. In addition, while short-term predictions (e.g., hourly) are often desired for building management purposes, the objective of this study requires prediction of the annual energy use of building by which the design team can assess energy performance of an abundance of building shape design alternatives at the early stage of the design process. Regarding the goal of assessments, the permissible error range would also be different from mentioned studies. Hence, there are major differences in goals, objectives, prediction inputs, accuracy and permissible error between this study and the previous studies on building energy use predictions using Deep Learning. The purpose of this paper is to assess the applicability of Deep Learning in predicting energy performance of a building based on its shape. In this regard, a DNN as the basic neural network architecture and a CNN for having potentials for facilitating communication within the design team are developed and assessed.

- **DNN**

Dense Neural Network (DNN), which is the simplest neural network architecture, is a biologically inspired computational model, designed to simulate the way in which the human brain processes information (Agatonovic-Kustrin and Beresford 2000). A DNN is composed of neurons that self-optimize through learning process. And as the name of DNN suggests, layers are densely connected, which means each neuron in a layer receives an input from all the neurons in the previous layer (i.e., each neuron is connected to all neurons of the previous layer) and similarly, it is an input for all the neurons in the following layers (Figure 2). DNNs can analyse complex data patterns and have been employed in many disciplines for a broad range of applications, from regression analysis and classification to un-supervised data clustering. In architectural design research, for instance, Turhan et al. (2014) use a DNN to predict heating load of a building based on the building aspect ratio, U-value of the envelope, area to volume ratio, and WWR.

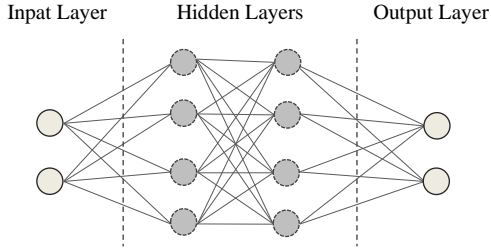

*Figure 2: Dense Neural Network.*

- **CNN**

A Convolutional Neural Network (CNN) is similar to a DNN, but is primarily developed to analyse visual imagery, which allows encoding image-specific features into the neural network architecture and making it more suited for image-focused tasks (O'Shea and Nash 2015). A CNN can map a large amount of data stored in an image to the final output through a simple yet precise architecture (Figure 3). In the past, DNN or multilayer perceptron models were employed for image recognition. However, the full connectivity between nodes caused the curse of dimensionality, whereas a CNN could substantially decrease the number of parameters due to having a weight sharing structure and pooling methods and therefore, outperform DNNs in analysing visionary images. It is important to note that CNN is spatially invariance, which means it does not encode the position and orientation of object. Hence, if the position of data matters, CNN may not be easily applicable. CNNs are increasingly being employed on a wide range of applications in architectural design research, from generative design to various classification tasks.

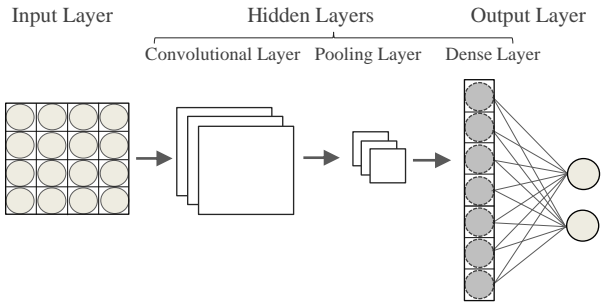

*Figure 3: Convolutional Neural Network.*

## Data Synthesis and Experiment Setup

To synthesize the dataset, a parametric model of a building is developed using Rhino-Grasshopper software tools. An existing building located on Texas A&M university campus is used as the initial model (Figure 4A). Table 2 summarizes the physical descriptions of the building model, which remain consistent through the experiment. It should be noted that the focus of this study is primarily to evaluate building shape design options at the early stages of the design. This is attained through controlling other influential factors such as the WWR, orientation, and thermal properties of the envelope to be unchanged, allowing for an informed decision on building shape design. Regarding that the emphasis of this study is on examining the applicability and comparison of neural networks in predicting building energy use based on the floor plan shape, for the sake of simplicity of the study, we limit the shape changes to certain structure, nonetheless, the parametric model generates a large search space. Each variation of the building shape can be quantified by four vectors; $X_1$, $X_2$, $X_3$, $X_4$ that change between -3.5m to 3.5m to keep the building shape reasonable. This also allows keeping the shape a semi-rectangle (excluding semi-cross shapes) and not imposing significant alteration in architectural layout plan. The area remains equal to the area of the rectangle a-b-c-d (Figure 4B).

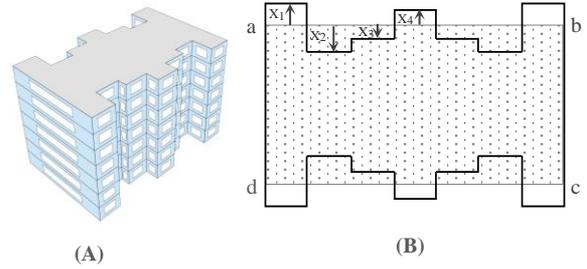

*Figure 4: (A) Building 3D model, (B) building shape can be quantified by four vectors; $X_1$, $X_2$, $X_3$, $X_4$.*

*Table 2: Building model attributes*

| Building program | Office: OpenOffice |
|---|---|
| Location | College Station, Texas |
| Number of floors | 7 |
| Area per floor | 990 m² |
| WWR | 30% |
| Width to length ratio | 0.5 |
| Orientation | 0° (Y- axis is in North direction) |
| Opaque construction | 100mm brick, 200mm heavyweight concrete, 50mm insulation board (Conductivity = 0.03 W/m-K), Wall air space resistance, 19mm gypsum board |
| Glazing construction | Clear 3mm, Air 13mm, Clear 3mm |
| Energy system | HVAC with Ideal Air Loads (Pre-defined by EnergyPlus) |
| Operation Schedule | Academic Calendar - office building |

It should be noted that a few dependent parameters, including building perimeter, total glazing and opaque surfaces area of the envelope, would inevitably change when building shape is changing, however, the WWR is kept constant, equal to 30% to minimize the changes in the total U-value of the envelope. Nevertheless, it is worth mentioning that a small change in façade solar transmittance is unavoidable. Moreover, the energy use and consequently the self-shading shape performance are dependent on climatic conditions, hence may vary in different locations.

The building operational energy is calculated through EnergyPlus as a tool for the whole building energy simulation, and the considered system boundary includes heating, cooling, and lighting loads. The experiment uses a controlled climate model. The TMY3 weather data of College Station, TX located in hot and humid climatic zone of the United States have been assigned to the model, to control the impact of climate on the model. The total annual energy required to sustain the interior space at the comfort temperature while meeting the typical office space loads i.e., occupancy, electrical equipment, lighting features, is calculated in kilowatt-hour. Comfort temperature and typical office space thermal loads are extracted from ASHRAE standard 55 and EnergyPlus schedule library, respectively.

To explore the applicability of neural networks in predicting building energy use based on its shape, two different Deep Learning structures; a DNN and a CNN are trained on the trainset and the prediction performance of the two models on the new samples stored as the test set are reported. The DNN model is fed with a list of four-vectors $(X_1, X_2, X_3, X_4)$, and the CNN model is fed with building floorplan (shape) images, saved in the black and white format and solid hatch as the input (Figure 5), and the energy simulation results to be trained for predicting the building energy use of new building designs. The image size was decided empirically by reducing the image size gradually up to 48×30 pixels, below which the network was adversely affected. It is worth mentioning that using grayscale images or other hatches also caused adverse effects.

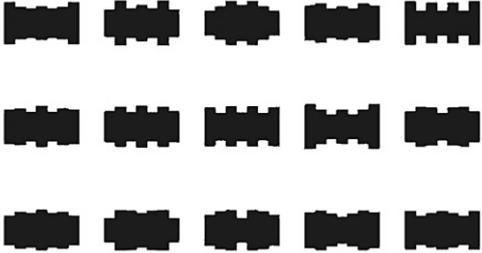

*Figure 5: A sample set of generated building shape images for feeding the CNN model.*

The number of hidden layers for the two networks are decided performing a grid search through a manually specified subset of the search space starting from the simplest architecture (i.e., a one-layer network), and considering the computation time per step. A selected number of the results of the grid search are displayed in Table 3 and Table 4 for the DNN and CNN models, respectively. The chosen number of hidden layers for the networks are shown in bold font, and the performance of the models are reported based on the prediction loss on the test set.

The DNN model consists of 64 dense layers, since for higher number of layers the loss was not decreasing considerably while the processing time was affected substantially, according to the tests. The CNN model consists of 32 convolutional layers. Similarly, for a higher number of layers, the CNN loss was not improving substantially. The training and testing are performed on a computer with an Intel (R) Core (TM) i7-7660U-2.50 GHz CPU, and 16 GB RAM. The models are developed using the Keras library, which provides a Python interface for Deep Learning. For the purpose of this study, it is important that the model does not predict far from the target, hence, the loss function is defined as the mean-squared-error (MSE) due to its high sensitivity to large errors compared to mean absolute error. Adam optimizer is chosen for the models' optimization process. The two models are trained with two dataset sizes, initially on a total of 350 samples and at the second experiment on a total of 1050 samples. The datasets are randomly divided into the train and test sets using a ratio of 80:20 for train: test. The chosen values for the hyperparameters are introduced in Table 5.

*Table 3: Number of layers in the DNN model*

| DNN Number of dense layers | Number of parameters | MSE | Time per step (s) |
|---|---|---|---|
| 2 | 7 | 1.268 | $28 \times 10^{-6}$ |
| 4 | 19 | 0.609 | $32 \times 10^{-6}$ |
| 8 | 43 | 0.155 | $46 \times 10^{-6}$ |
| 16 | 91 | 0.061 | $43 \times 10^{-6}$ |
| 32 | 187 | 0.035 | $36 \times 10^{-6}$ |
| **64** | **379** | **0.019** | **$46 \times 10^{-6}$** |
| 128 | 763 | 0.011 | $43 \times 10^{-6}$ |
| 256 | 1531 | 0.008 | $46 \times 10^{-6}$ |
| 512 | 3067 | 0.006 | $46 \times 10^{-6}$ |

*Table 4: Number of layers in the CNN model*

| CNN Number of conv layers | Number of parameters | MSE | Time per step (s) |
|---|---|---|---|
| 2 | 665 | 0.711 | $271 \times 10^{-6}$ |
| 4 | 1329 | 0.692 | $556 \times 10^{-6}$ |
| 8 | 2657 | 0.643 | $296 \times 10^{-6}$ |
| 16 | 5313 | 0.627 | $363 \times 10^{-6}$ |
| **32** | **10625** | **0.430** | **$602 \times 10^{-6}$** |
| 64 | 21249 | 0.552 | $10^{-3}$ |
| 128 | 42497 | 0.554 | 1.002 |
| 256 | 84993 | 0.576 | 1.004 |

*Table 5: Developed Models Description*

|  | CNN | DNN |
|---|---|---|
| Input shape | (n, 48, 30) | (n, 4) |
| Number of layers | Total of 34 | Total of 64 |
| Batch size | 32 | 32 |
| Number of epochs | 100 | 100 |
| Learning rate | $10^{-3}$ | $10^{-3}$ |

## Results and Discussion

As was discussed earlier, this study pursues two goals; firstly, to evaluate the applicability of neural networks in building energy use prediction based on the building shape; secondly, to compare the two neural networks in terms of performance, simplicity, and computation time.

Based on the results of the first experiment with the smaller dataset, the total annual energy use of building is predicted with an estimated root-mean-squared-error of 0.14 and 0.66 using the DNN and CNN model, respectively. The comparison between the two neural networks shows that the DNN model surpasses the CNN model in performance, simplicity, and computation time. The DNN model predicts the annual energy use of a building based on its shape within an estimated MSE of 0.019. These results can be improved by increasing the depth of the model. For instance, by increasing the number of layers from 64 to 512 layers, the model achieves an estimated MSE of 0.006. However, for the purpose of comparison, according to the efficiency of the models the number of layers for the DNN model is kept equal to 64.

The CNN model predicts the building energy use with an estimated MSE of 0.430. Figure 6A shows the energy simulation results versus the CNN model predictions on the test samples. The dotted line shows the simulation results, and the solid blue line shows the CNN model predictions. The CNN model has a total of 34 layers consisting of 32 convolutional layers (with filter size of 3×3), a max pooling of a 2×2 size, and a dense layer. This structure of the CNN model results in approximately 28 times as many parameters as in the DNN model. Similarly, the computation time per step is longer for the CNN model and is approximately 13 times as much as the DNN model takes per step (Table 6). It should be noted that the results are an estimation and not a determined value given the stochastic nature of the models and the gradient descent initialization. The CNN prediction does not show further significant improvement by increasing the depth of model based on the test experiments. Moreover, on this experiment with smaller dataset size, use of cross validation does not lead to a considerable improvement on the results, which may be justified by the limited trainset size.

*Table 6: DNN and CNN model comparison*

| Model | Dataset size | Number of parameters | Number of epochs | Batch size | MSE |
|---|---|---|---|---|---|
| **DNN** | 350 | 379 | 100 | 32 | 0.019 |
| **CNN** | 350 | 10,625 | 100 | 32 | 0.430 |

At the next stage, the dataset size is increased to 1050 samples, which is 3 times as many as in the first experiment with the smaller dataset. The results show that both neural networks are thoroughly capable of predicting a building operational energy use based on its shape, once trained. The CNN results significantly improve, when the dataset is increased, as can be seen in Figure 6B. The dotted line shows the EnergyPlus simulation results, and the solid blue line shows the CNN model predictions on the test samples. On the second experiment with larger dataset, the CNN model performs considerably better, but still does not surpass the DNN model. It can be concluded that the first limitation of a CNN model is the requirement of a larger dataset for training due to having a larger number of parameters. The DNN model yields an estimated MSE of 0.008, which is much lower than the MSE resulted by the CNN model (approximately 0.022) after employing cross-validation in the training process.

It is notable to find out that increasing the size of the dataset shows a higher effect on the CNN model. Moreover, employing cross-validation significantly improves the prediction results in the CNN model, when the dataset size was increased. Increasing the size of the dataset results in a decrease of 50% in MSE of the DNN model, and a decrease of 95% in MSE of the CNN model on the test set, compared to the initial results with 350 samples. Both models, the CNN and the DNN, performs well, when the dataset size is increased, and cross-validation is employed in the training process (Table 7). Comparing the two models in terms of complexity, the CNN model has considerably higher number of parameters (equal to 10,625 parameters), while the DNN number of parameters is 379. To reduce the number of parameters of the CNN model, the filters are enlarged to 5×5-pixel filters and the pooling size is increased to 4×4, by which the model's number of parameters is reduced to 2,945 while it did not result in a significant decrease in the model's performance.

**The CNN model**

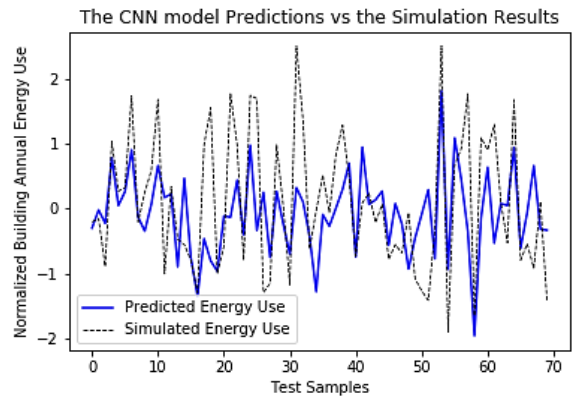

(A)

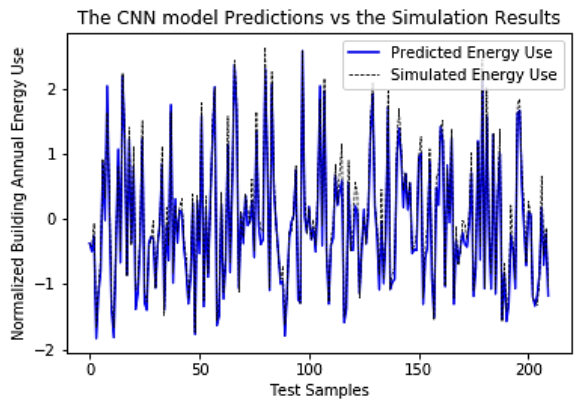

(B)

*Figure 6: EnergyPlus simulation results versus the CNN model predictions for building energy use (A): with smaller dataset size (350 samples), (B): with larger dataset size (1050 samples).*

*Table 7: CNN and DNN model comparison*

| Model | Dataset size | Number of parameters | Number of epochs | Batch size | MSE |
|---|---|---|---|---|---|
| **DNN** | 1050 | 379 | 100 | 32 | 0.008 |
| **CNN** | 1050 | 10,625 | 100 | 32 | 0.022 |

## Conclusion and Future Work

This study uses Deep Learning models for fast prediction of building energy use based on its shape, to achieve energy efficient building design at the early stage of the design process. In this regard, a DNN as the basic neural network architecture and a CNN for having potentials for facilitating communication within the design team are trained with two different training set sizes and the prediction performance of the two models on the test sets are compared in terms of performance, simplicity, and computation time. A synthesized dataset, using a parametric building model developed in Rhino-Grasshopper, and the corresponding building operational energy use calculated using EnergyPlus is utilized to train and test the models. The analysis of the results achieved by comparing models shows that both neural networks are thoroughly capable of predicting energy use of a building based on its shape, once trained. The DNN performs faster and yields significantly lower loss (MSE). The CNN performance improves considerably, when the dataset size is increased, but still does not surpass the DNN model. Moreover, the CNN model has a higher number of parameters, which increases the computation time of the model to approximately 602 microseconds per step while the computation time for the DNN model is approximately 46 microseconds per step. Hence, two of the important challenges of the CNN model are the requirements of a larger amount of training data and a longer computation time. Nevertheless, due to the use of graphical design documents that are produced in the design process and extensively utilized in design communication between architects and engineers, a CNN model may better facilitate communication than a DNN model within the design team through using drawings and graphical representations of the design options. The graphical data input for CNN can potentially be extracted from any of the building models created in the design process (e.g., BIM or CAD). This potential motivates us for further studies on exploring ways to improve the prediction performance and creative use of CNN, in addition to exploring ensemble methods considering both methods' inherent strengths. Moreover, there are several aspects of the developed models (e.g., generalizability, and efficiency) that require further studies.